\documentclass[runningheads]{llncs}
\usepackage{graphicx}
\usepackage{amsmath,amssymb} % define this before the line numbering.
\usepackage{color}
\usepackage{booktabs} % For formal tables
\usepackage{xspace}
\usepackage{subfigure}
\usepackage[skip=0pt]{caption}
\begin{document}
\pagestyle{headings}
\mainmatter

\def\ACCV20SubNumber{}  % Insert your submission number here

%===========================================================
\title{Image Retrieval for Structure-from-Motion via Graph Convolutional Network} % Replace with your title
\titlerunning{ACCV-20 submission ID \ACCV20SubNumber}
\authorrunning{ACCV-20 submission ID \ACCV20SubNumber}

\author{Shen Yan, Yang Pen, Shiming Lai, Yu Liu, Maojun Zhang}
\institute{Paper ID \ACCV20SubNumber}

\maketitle

%===========================================================
\begin{abstract}
Conventional image retrieval techniques for Structure-from-Motion (SfM) suffer from the limit of effectively recognizing repetitive patterns and cannot guarantee to create just enough match pairs with high precision and high recall. In this paper, we present a novel retrieval method based on Graph Convolutional Network (GCN) to generate accurate pairwise matches without costly redundancy. We formulate image retrieval task as a node binary classification problem in graph data: a node is marked as positive if it shares the scene overlaps with the query image. The key idea is that we find that the local context in feature space around a query image contains rich information about the matchable relation between this image and its neighbors. By constructing a subgraph surrounding the query image as input data, we adopt a learnable GCN to exploit whether nodes in the subgraph have overlapping regions with the query photograph. Experiments demonstrate that our method performs remarkably well on the challenging dataset of highly ambiguous and duplicated scenes. Besides, compared with state-of-the-art matchable retrieval methods, the proposed approach significantly reduces useless attempted matches without sacrificing the accuracy and completeness of reconstruction.  

\end{abstract}

%===========================================================
\section{Introduction}
\label{sec:intro}

%-------------------------------------------------------------------------
% what the problem is?
Contemporary Structure-from-Motion (SfM) systems~\cite{agarwal2011building,moulon2013global,schonberger2016structure} widely employ image retrieval techniques to relieve the heavy computational burden of image matching process, assuming that image pairs only with high visual similarity are likely to match. The retrieval methods for SfM are commonly implemented within two steps: Step 1, map every image in the dataset to individual vectors via an embedding function; Step 2, for each query image, find its nearest neighbors through a certain similarity metrics between quantized vectors. A variety of approaches have been developed in such two operations. For example, in Step 1, vocabulary tree models~\cite{sivic2003video,nister2006scalable} or CNN-based approaches~\cite{radenovic2016cnn,shen2018matchable} are proposed to describe image features as a whole. In Step 2, KD-Tree~\cite{bentley1975multidimensional} or Ball-Tree~\cite{omohundro1989five} is often adopted to accelerate the approximate search. However, while the index techniques have shown promising results in effectively filtering unnecessary matches, they have still been underachieving.

\begin{figure}
	\centering
	\includegraphics[clip,trim=0mm 40mm 0mm 0mm,width=0.98\textwidth]{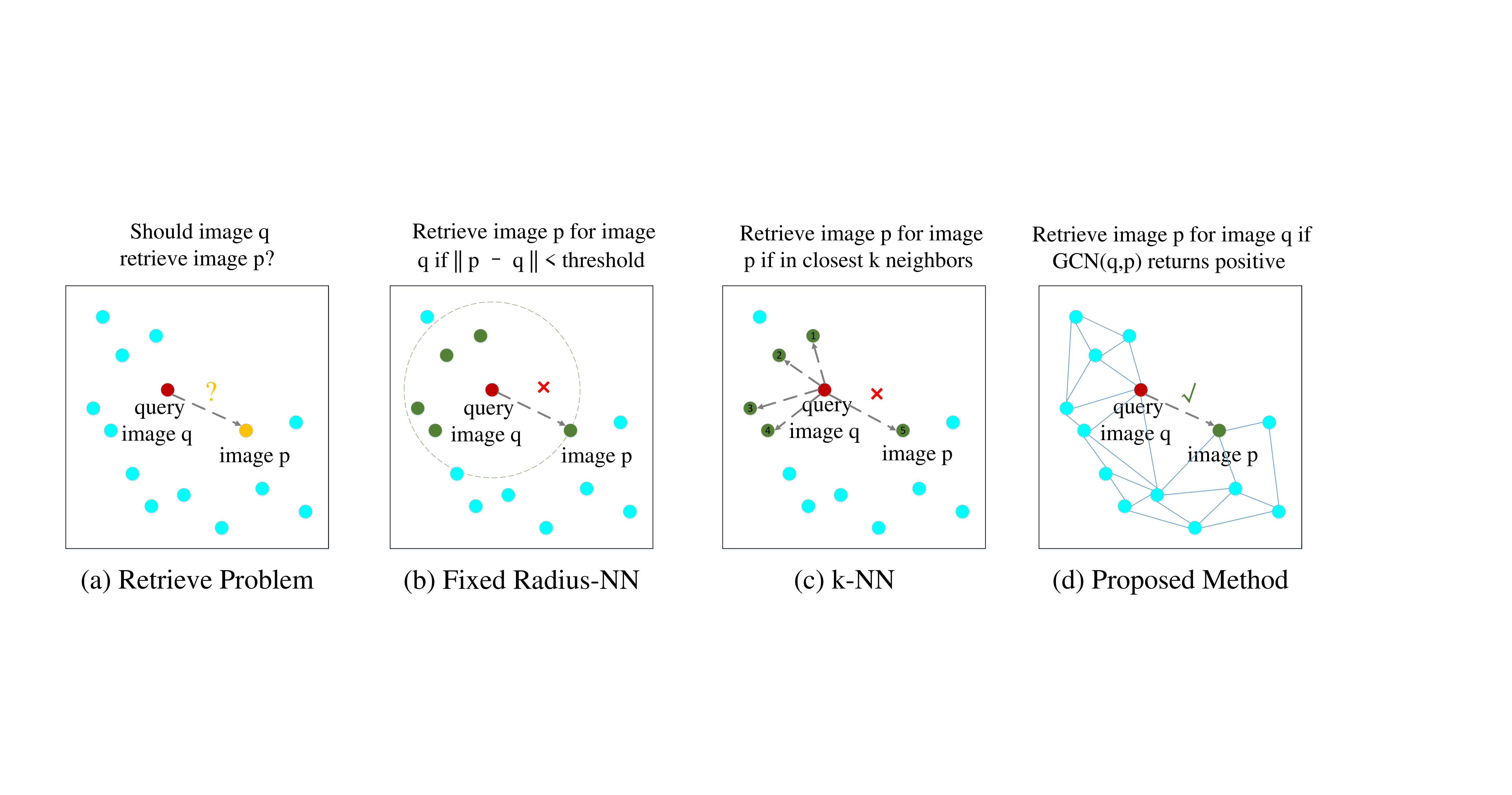}
	%\goodgap
	\caption{Basic idea of our method. (a) This paper intends to perform image retrieval. (b-c) Directly retrieve image with hyper-parameters ($\tau$ or $k$). (d) Our idea: use GCN to learn surrounding local context for retrieval prediction.}
	\label{fig:idea}
%	\vspace{-0.78cm}
\end{figure}

%------------------------------------------------------------------------- 
% the best work face what kind of challenge?
We believe that former retrieval techniques face two major challenges. First, the embedding function in Step 1 is vulnerable to symmetric or repetitive textured patterns in ambiguous scenes. The embedding function, either trained by vocabulary tree or CNN model, cannot make features extracted from ambiguous structure manifest a notable difference. Such visually similar yet distinctive patterns are therefore inappropriately identified as overlapping in Step 2. To make matters worse, these misidentified pairs not only deceive the retrieval algorithms but can also pass the two-view geometry verification and form erroneous pairwise epipolar geometry. The false matches significantly mislead the direction of reconstruction and give rise to incomplete folded structure or total collapse of SfM. As a result, a sufficient subset of matches that does not contain potentially wrong epipolar geometry is superior to redundant matches that may have incorrect matching pairs for 3D reconstruction.
  
Second, it is extremely difficult to set up exactly sufficient match pairs for SfM. Previous research usually tries to achieve this purpose by empirically adjusting the number of retrieved items $k$ or the similarity threshold score $\tau$ in Step 2. The problem is that, on the one hand, smaller hyper-parameter ($k$ or $\tau$) settings will cause the missing of true positive matches, which may lead to a decline in the completeness of SfM or even model disconnections. On the other hand, larger hyper-parameter ($k$ or $\tau$) settings will bring about some false positive matches, which will inevitably result in inefficiency and inaccuracy of SfM. Moreover, as the density of scenic spots varies, it is improper to assume that all views share a consistent amount of index items $k$.

%------------------------------------------------------------------------- 
% our work can release the mentioned challenges
To mitigate the two challenges mentioned above, we introduce a novel retrieval method based on Graph Convolutional Network (GCN) to generate accurate pairwise matches without costly redundancy. The framework of the proposed method can be summarized as follow.

For a query image, we build a Query Enclosing Subgraph (QES) around it to bring in candidate retrieved items and depict its local context. The motivation behind this work is that the similarity likelihood between a node and its neighbors can be effectively reasoned from its local topological information~\cite{zhang2017weisfeiler,zhang2018link,wang2019linkage}. Then, we adopt a Graph Convolutional Network (GCN) to learn to integrate valuable context knowledge and classify nodes in QES with positive or negative output label. All positive samples are regarded as sharing scene overlaps with the query image. Note that we only consider nearest neighbors of the query image as candidate retrieved items in practice, as only a few of matches are needed to be retrieved in SfM. 
  
Since our algorithm grasps context information provided by QES, the measure of similarity which is unable to be computed in image feature space, can be effectively calculated in topological space. The symmetric and repetitive textured patterns can be successfully distinguished as they reveal different properties in the later space. Besides, due to the fact that GCN model directly returns prediction results for candidate index images, it is no longer necessary to try to carefully select hyper-parameters ($k$ or $\tau$). We show that the retrieved items inferred from GCN model basically cover all required match pairs for SfM and do not contain too much redundancy. The main idea of the proposed method is illuminated in Fig~\ref{fig:idea}.

%------------------------------------------------------------------------- 
% our main contribution
Our main contributions could be summarized as follows:

\begin{enumerate}

\item We convert the problem of matchable image retrieval to the problem of node binary classification in subgraphs, which helps to overcome scene ambiguous difficulty.

\item We propose a learnable GCN to automatically predict the matchable relationship between candidate image pairs, and the generated pairwise matches are proved to be exactly enough for SfM.

\item We conduct extensive experiments on various kinds of 3D reconstruction datasets and compare our approach to vocabulary tree and CNN-based models. Our method outperforms state-of-the-art retrieval methods on challenging ambiguous dataset and can offer precisely enough matchable pairs for SfM. 

\end{enumerate}

%------------------------------------------------------------------------- 

%===========================================================
\section{Related Work}
\label{sec:related}

%------------------------------------------------------------------------- 
\subsection{Image Retrieval Techniques for Structure-from-Motion}

% voc tree methods

%------------------------------------------------------------------------- 

Vocabulary tree~\cite{sivic2003video,nister2006scalable} is the most extensively used technique to rank images in dataset given a query photo, which has been implemented by most publicly available SfM pipelines~\cite{agarwal2011building,moulon2013global,sweeney2015optimizing,schonberger2016structure,shen2016graph} as a preemptive pruning step. A vocabulary tree is learned typically from hierarchically clustering local feature descriptors of all images in dataset. Then, Term Frequency Inverse Document Frequency (TF-IDF) is utilized to efficiently score the similarity of images with inverted files. Research community reaches a higher level of maturity by improving quantization procedure~\cite{philbin2007object,jegou2008hamming,philbin2008lost}, adopting compact representations~\cite{perronnin2010large,jegou2011aggregating,radenovic2015multiple,arandjelovic2013all,tolias2014orientation}, incorporating geometric cues~\cite{chum2009large,jegou2008hamming,philbin2007object,shen2013spatially}, and applying query expansion~\cite{chum2011total,qin2011hello,tolias2014visual}. Although vocabulary tree assists SfM pipelines to eliminate computation cost, substantial memory footprints are still required during both constructing and indexing processes.

% CNN-based methods

%------------------------------------------------------------------------- 

Recent developments~\cite{babenko2014neural,sharif2015baseline,tolias2015particular,babenko2015aggregating,babenko2015aggregating} illuminate that Convolutional Neural Networks (CNN) offer an attractive alternative for image encoding with small memory footprint. Object retrieval task has already applied deep CNN descriptors to represent images. However, most CNN-based object retrieval methods build on the assumption that images should share salient semantic regions like landscapes or architecture. In real 3D reconstruction, many photographs merely serve as bridge to connect partial scenes, with discontinuous or even no semantically meaningful regions.

Filip Radenovic et al.~\cite{radenovic2016cnn} and Shen et al.~\cite{shen2018matchable} specialize on solving the matchable image retrieval mission of 3D reconstruction. They employ state-of-the-art reconstruction algorithms to rebuild 3D models, which are re-projected on images to generate ground-truth supervised data. These training data ensure that images will be retrieved according to scene overlaps rather than semantic similarity. Filip Radenovic et al. adopt a siamese architecture with contrastive loss~\cite{chopra2005learning}. Besides, they introduce learned whitening and R-MAC~\cite{tolias2015particular} to improve performance. Shen et al. employ a triplet loss~\cite{wang2014learning,schroff2015facenet} architecture, with pre-matching regional code (PRC) to boost accuracy at the expense of reducing efficiency.

% retrieve methods comments

%------------------------------------------------------------------------- 
 In summary, matchable image retrieval task in 3D reconstruction have experienced developments from vocabulary tree to CNN-based methods. The most important part in previous work is to find an embedding function to map images into a compact feature space. However, these methods only consider visual information, leading to the result that ambiguous patterns cannot be convincingly differentiated. Besides, previous research ignores how to acquire exactly enough match pairs for SfM.

\subsection{Graph Convolutional Network (GCN)}

% GCN introduction
Recently, there is increasing interest in extending deep learning approaches for graph data~\cite{bruna2013spectral,defferrard2016convolutional,kipf2016semi,hechtlinger2017generalization,hamilton2017inductive,velivckovic2017graph,yan2018spatial}, such as e-commence, social network and molecular chemistry. Similar to using CNN on Euclidean data, GCN is proposed to deal with irregular graph data. According to the definition of convolution on graph structure, GCN is divided into two main streams, the spectral-based approaches~\cite{bruna2013spectral,defferrard2016convolutional,kipf2016semi} and the spatial-based approaches~\cite{hechtlinger2017generalization,hamilton2017inductive,velivckovic2017graph,yan2018spatial}. Spectral-based GCN develops a graph convolution based on Graph Fourier Transform theory, while spatial-based GCN directly performs manually-defined convolution based on a node’s spatial relations.

GCN has many applications across different tasks and domains including node classification~\cite{shchur2018pitfalls} and link prediction~\cite{liben2007link}. Actually, link prediction could be regarded as binary classification problem. Traditional methods calculate the linkage likelihood between two given nodes by developing carefully designed heuristics~\cite{brin1998anatomy,jeh2002simrank,zhou2009predicting,katz1953new}. However, a significant limitation of these heuristics is that they lack universal applicability to different kinds of graphs. Zhang and Chen therefore propose a Weisfeiler-Lehman Neural Machine~\cite{zhang2017weisfeiler} and a graph neural network~\cite{zhang2018link} to learn general subgraph structure for linkage likelihood computation. Based on their work, Wang et al. further propose a linkage based face clustering algorithm~\cite{wang2019linkage}, utilizing potential identities to group a set of faces. These methods are closely related to our work, since we solve the image retrieval problem by adopting a GCN to infer the matchable information between a query image and its neighbors.

%------------------------------------------------------------------------- 

\section{Proposed Approach}
\subsection{Overview}
\label{3.1}
Assume that we have a collection of $N$ unordered images $\lbrace I_1, \cdots, I_q \cdots, I_N \rbrace$ with geometric overlaps, for each query image $I_q$, we aim to find an index set $S^q= \lbrace  I^q_1,I^q_2,...,I^q_k \rbrace$, where $k$ is the number of retrieved items. To find the retrieval set, one typical pipeline is to first map images $\lbrace I_1, \cdots, I_q \cdots, I_N \rbrace$ into a certain compact feature space via an embedding function $f(\cdot)$. Then nearest neighbors of $I_q$ is searched by a defined similarity measurement $D(f(I_q),f(I_p))$.

\begin{equation}
D(f(I_q),f(I_p)) = \left\|{\frac{f(I_q)}{\left\|{f(I_q)}\right\|}-\frac{f(I_p)}{\left\|{f(I_p)}\right\|}}\right\|_2.
\label{equ:distance}
\end{equation}

\begin{figure}
	\centering
	\includegraphics[clip,trim=0mm 20mm 0mm 0mm,width=0.98\textwidth]{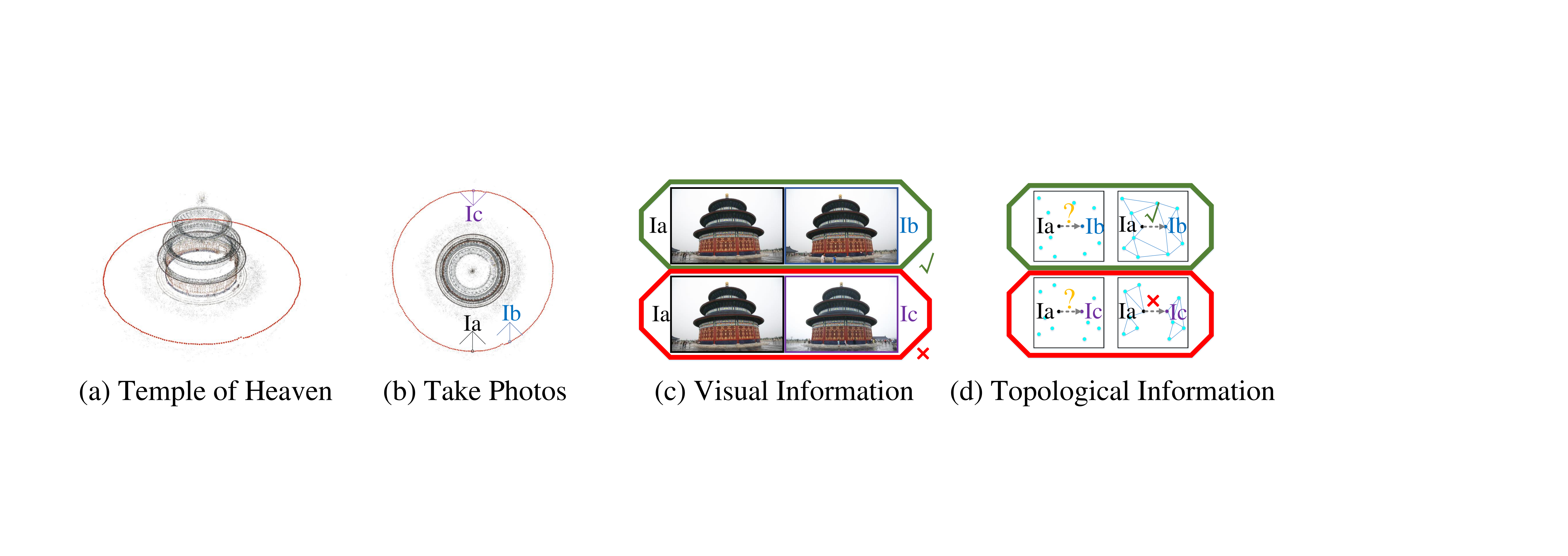}
	%\goodgap
	\caption{Example of Temple-of-Heaven dataset.}
	\label{fig:method}
%	\vspace{-0.78cm}
\end{figure}

However, when images are acquired from a site with highly ambiguous structure, this methodology fails. In what follows we provide an example for illustration. Fig~\ref{fig:method}(a) shows an extremely symmetric dataset Temple-Of-Heaven, which is composed of 341 rotationally symmetric images. In Fig~\ref{fig:method}(b-c), we observe that some photos taken in quite different positions look extremely similar. This phenomenon explains why it is unreliable to retrieve images only by visual features. Fortunately, in Fig~\ref{fig:method}(d), we notice that though two image pairs both staying close in feature space, they have totally different distances in topological space. Intuitively, we consider utilizing the local context surroundings to supply extra information for boosting matchable retrieval performance.

 Suppose that the image collections $\lbrace I_1, \cdots, I_q \cdots, I_N \rbrace$ have already been embedded as $\lbrace f(I_1), \cdots, f(I_q) \cdots, f(I_N) \rbrace$, we treat each feature $f(I_i)$ as node $i$. For query node $q$, the primary action is to exploit some kind of data structure to describe its local context. We look on query node $q$ as a center, and build a subgraph around it called Query Enclosing Subgraph (QES). The construction of QES is described in detail in Section~\ref{3.2}. Given a QES as input data, we then deploy a Graph Convolution Network (GCN) on it for node binary classification and the network directly outputs retrieved items (marked as positive). The mechanism of GCN is presented in Section~\ref{3.3}.

\subsection{Construction of Query Enclosing Subgraph}
\label{3.2}
\begin{figure}
	\centering
	\includegraphics[clip,trim=0mm 20mm 0mm 0mm,width=0.98\textwidth]{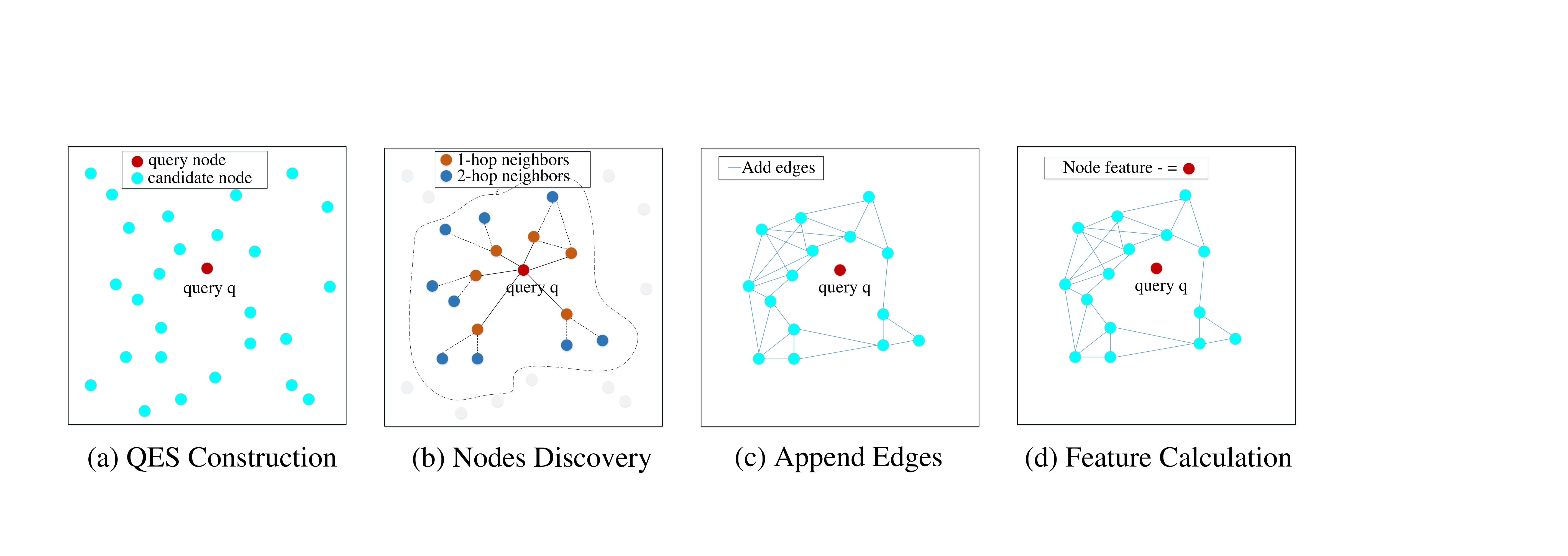}
	%\goodgap
	\caption{Construction of Query Enclosing Subgraph (QES) with three stages.}
	\label{fig:QES}
%	\vspace{-0.78cm}
\end{figure}

For query node $q$, Query Enclosing Subgraph (QES) is represented as $G^{q} = (V^{q},E^{q})$, where $V^{q}$ is the set of nodes, and $E^{q}$ is the set of undirected edges. Let $v^q_i \in V^q$ denote a node and $e^q_{ij} = (v^q_i,v^q_j) \in E^q$  denote an undirected edge between $v^q_i$ and $v^q_j$, and $n$ represents the number of nodes. The adjacency matrix $\mathbf{A}^{q}$ is an $n \times n$ matrix with $A^q_{ij} = 1$ if $e^q_{ij} \in E^{q}$ and $A^q_{ij} = 0$ if $e^q_{ij} \notin E^{q}$. $G^{q}$ has node attributes $\mathbf{X}^{q}$, where $\mathbf{X}^{q} \in \mathbf{R}^{n \times d}$ is a node feature matrix with $\mathbf{x}_{v^q} \in \mathbf{R}^d$ indicating the feature vector of a node $v^q$. As QES consists of three different data types, namely nodes $V^q$, edges $E^q$ and features $\mathbf{X}^q$, we correspondingly construct QES by three stages illustrated in Fig~\ref{fig:QES}.

\subsubsection{Stage 1: Node discovery}
We use $\lbrace N_k({i}) \rbrace$ to denote the set of $k$ nearest neighbors ($k$NNs) of a node $i$, which are searched through Equation~\ref{equ:distance}. For query node $q$, we first add its 1-hop $k$NNs nodes $\lbrace N_{k_1}({q}) \rbrace$ to the unordered node list $V^{q}$. Then, nodes $\lbrace N_{k_2}(p) | p \in N_{k_1}(q) \rbrace$ in 2-hop are iteratively added to the node set  $V^{q}$.  $k_1, k_2$  denote the number of nearest neighbors in the first and second hop respectively. Although this chain can be continuously extended, we only sample $k$NNs of $q$ up to 2-hop. This is because 2-hop QES already covers all the information needed to calculate any first and second-order heuristics for link prediction.  In addition, note that query node $q$ itself is excluded from $V^q$.  

\subsubsection{Stage 2: Append edges among nodes}
Assuming we have obtained a node set $V^q$ from Stage 1, the next step is appending edges among the nodes. We traverse all nodes $\lbrace p \in V^q \rbrace$, search its $u$NNs $\lbrace N_u(p)\rbrace $ among all nodes in original entire dataset. If a node $r \in \lbrace N_u(p)\rbrace $ also appears in $V^q$, we insert an undirected edge $e^q_{pr}$ into the edge set $E^q$.

\subsubsection{Stage 3: Node feature calculation}
The embedding function $f(\cdot)$ used to extract image global feature is a pre-trained CNN-based~\cite{shen2018matchable} model. We assign each node $v^q \in V^q$ with extracted feature vector $\mathbf{x}_{v^q} = f(I_{v^q})$. In order to make $v^q \in V^q$ share information of query node $q$, we uniformly subtract node features $\lbrace \mathbf{x}_{v^q} \rbrace$ by query feature $\mathbf{x}^{q} = f(I_q)$. The final feature matrix $\mathbf{X}^{q}$ is described as follows:

\begin{equation}
\mathbf{X}^{q} = {\lbrack \cdots, \mathbf{x}_{v^q} - \mathbf{x}_{q} \cdots \rbrack}^T, \ for \ all \ v^q \in V^q.
\label{equ:normalize}
\end{equation}

\begin{figure}
	\centering
	\includegraphics[clip,trim=0mm 0mm 0mm 0mm,width=0.9\textwidth]{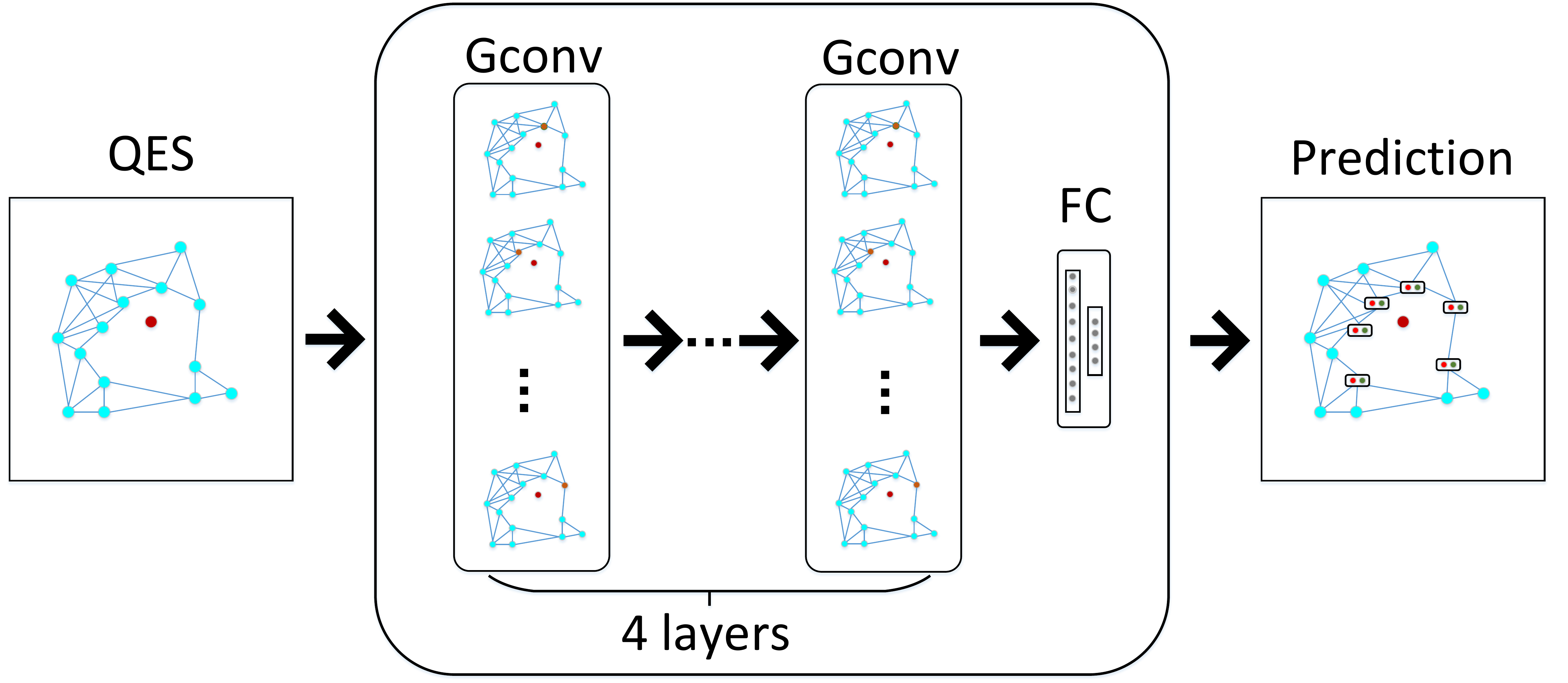}
	%\goodgap
	\caption{Overall architecture of GCN.}
	\label{fig:GCN}
%	\vspace{-0.78cm}
\end{figure}

\subsection{Graph Convolutional Network on QES}
\label{3.3}
After accomplishing the construction of QES for query node $q$, we apply a Graph Convolution Network (GCN) on it to perform retrieving. The GCN determines if a node $v^q$ in QES is positive (should be retrieved to query node $q$) or negative (should not be retrieved to query node $q$). Specifically, we introduce the adopted GCN in two aspects: the graph convolutional layer and the overall architecture.

\subsubsection{Graph Convolutional Layer} 
The graph convolutional layer basically follows GCN~\cite{kipf2016semi} with slight modifications, who takes node feature matrix $\mathbf{X}^{q}$ together with an adjacency matrix $\mathbf{A}^{q}$ as input and output a filtered feature matrix $\mathbf{Y}^{q}$. 

A graph convolutional layer first encapsulates each node’s hidden representation by aggregating feature information from its neighbors. This operation is achieved by left multiplying $\mathbf{X}^{q}$ by an aggregation matrix $\mathbf{G}^q$. The aggregation matrix is defined as $\mathbf{G}^q = {\mathbf{\Lambda}^q}^{-\frac{1}{2}} \mathbf{A}^q {\mathbf{\Lambda}^q}^{-\frac{1}{2}}	$, where $\mathbf{\Lambda}^q $ is a diagonal matrix with $\mathbf{\Lambda}^q_{ii} = \sum_j \mathbf{A}^q_{ij} $. Then we concatenate feature matrix $\mathbf{X}^{q}$ with aggregation feature matrix $\mathbf{G}^q \mathbf{X}^{q}$ along the feature dimension. After feature aggregation, a non-linear transformation is applied to the resulted outputs, the weight matrix parameters $\mathbf{W}^q$ is to be learned. Formally, a graph convolutional layer in our paper has the following formulation,

\begin{equation}
\mathbf{Y}^q = \sigma ( \lbrack \mathbf{X}^q || \mathbf{G}^q \mathbf{X}^q \rbrack \mathbf{W}^q ),
\label{equ:gcn}
\end{equation}
where operator $||$ represents matrix concatenation and $\sigma( \cdot )$ is a non-linear activation function. 

\subsubsection{Overall Architecture}

The proposed GCN model can be regarded as a combination of two components: feature extraction part and the node classification part, as shown in Fig~\ref{fig:GCN}. For feature extraction part, the main block is a stack of four graph convolution layers activated by the ReLU function. By stacking four layers, the final hidden representation of each node receives messages from a further neighborhood. After that, we add a couple of fully connected layers in order to wrap up the high-level node representations. For node classification part, we use the cross-entropy loss function after the sigmoid activation for optimization. Because only a few of retrieved items matter in SfM, in train phase, we only backpropagate the gradient for the nodes of the 1-hop neighbors; in test phase, we perform node binary classification on the 1-hop nodes as well. 

\begin{figure}
	\centering
	\subfigure[View graph of CNN-based method]{
		\label{fig:effect 1}
		\includegraphics[clip,trim=0mm 0mm 0mm 0mm,width=0.46\textwidth]{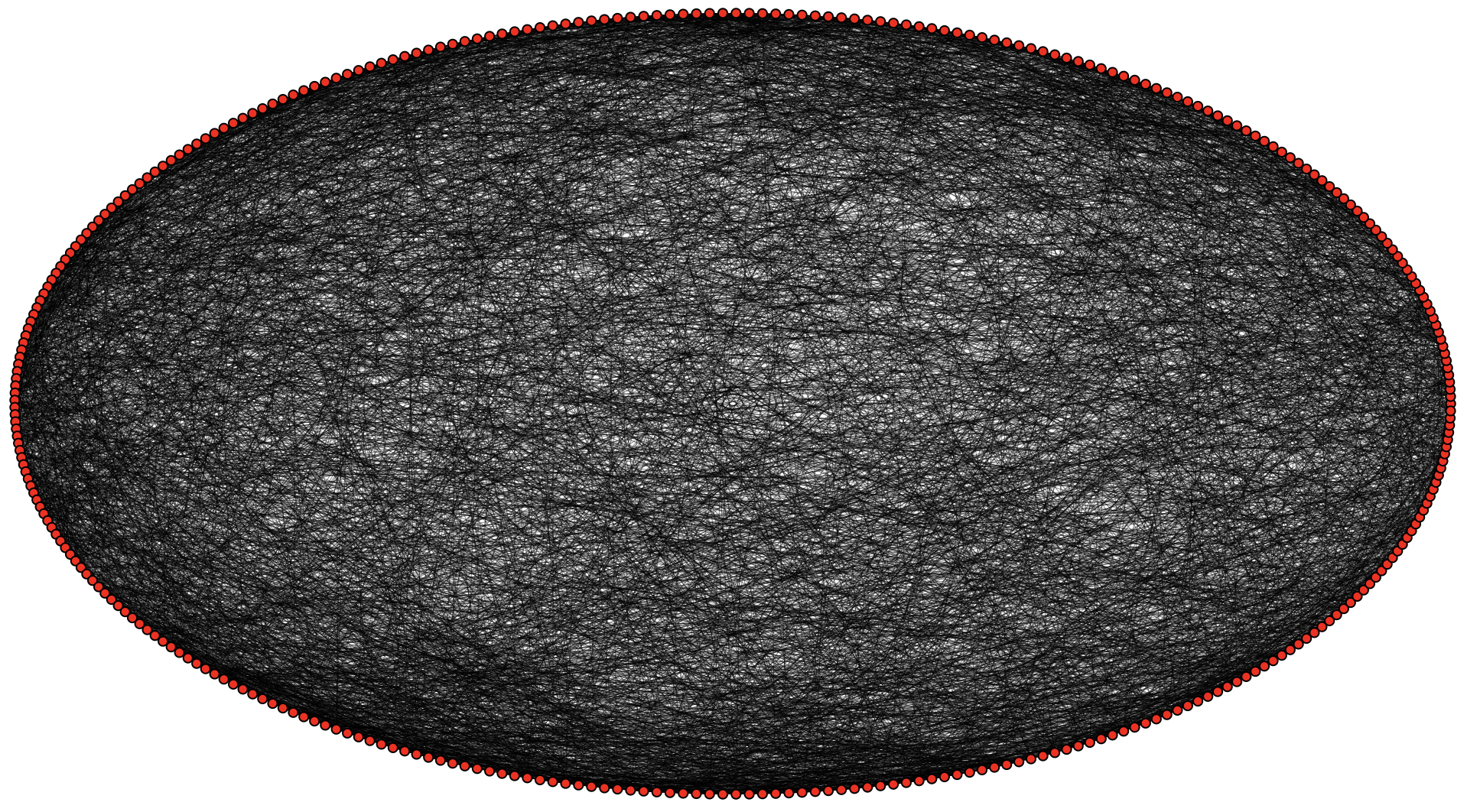}
	}
	%\goodgap
	\subfigure[View graph of our method]{
		\label{fig:effect 2}
		\includegraphics[clip,trim=0mm 0mm 0mm 0mm,width=0.46\textwidth]{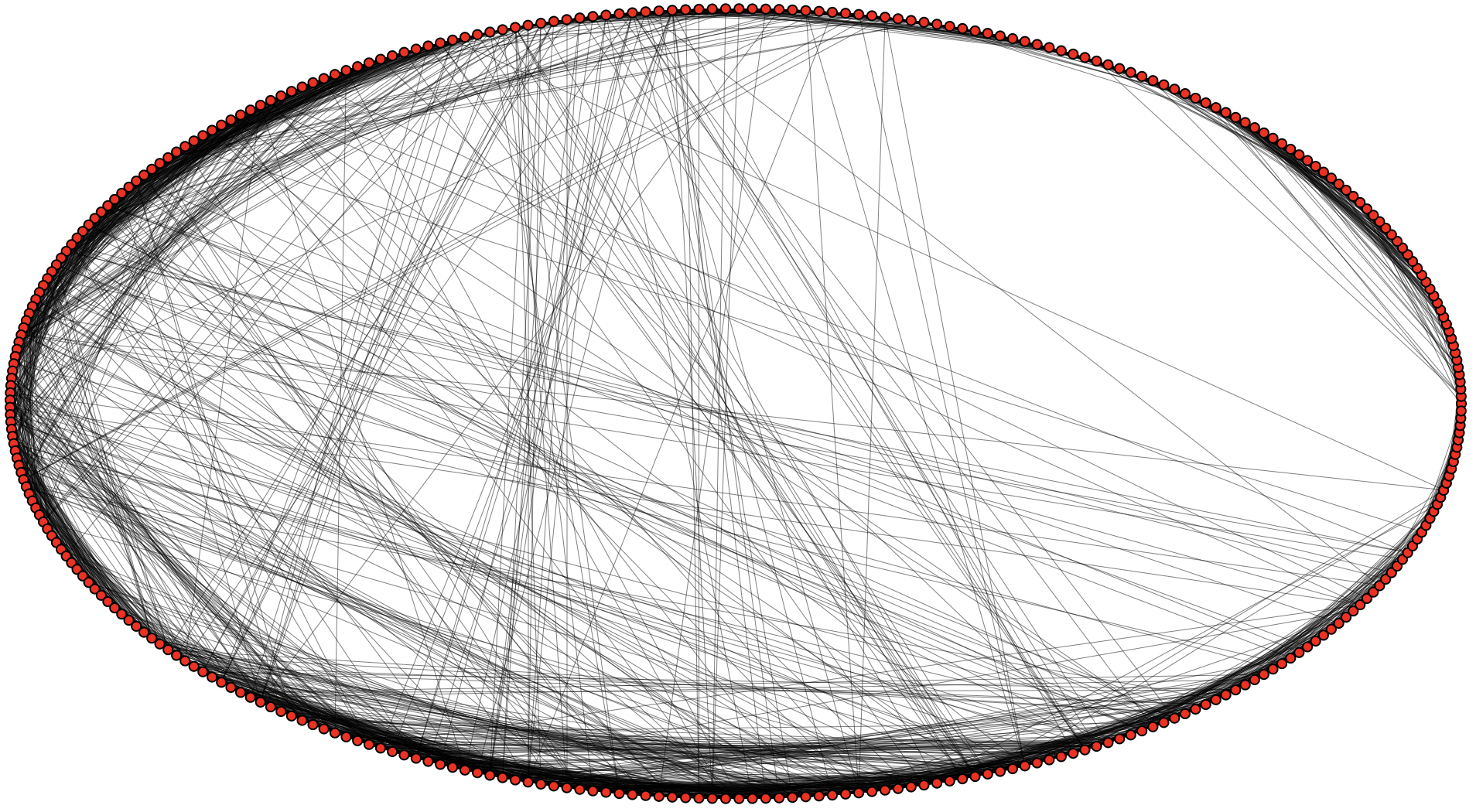}
	}
%    \vspace{-0.4cm}
	\caption{(a) We observe that there are numerous match errors by directly applying pretrained CNN-based models with $k=25$. (b) The proposed GCN method effectively eliminates these mistakes and finds sufficient retrieved images.}
	\label{fig:effect}
%	\vspace{-0.78cm}
\end{figure}

\begin{figure}
	\centering
	\includegraphics[clip,trim=0mm 15mm 0mm 10mm,width=0.98\textwidth]{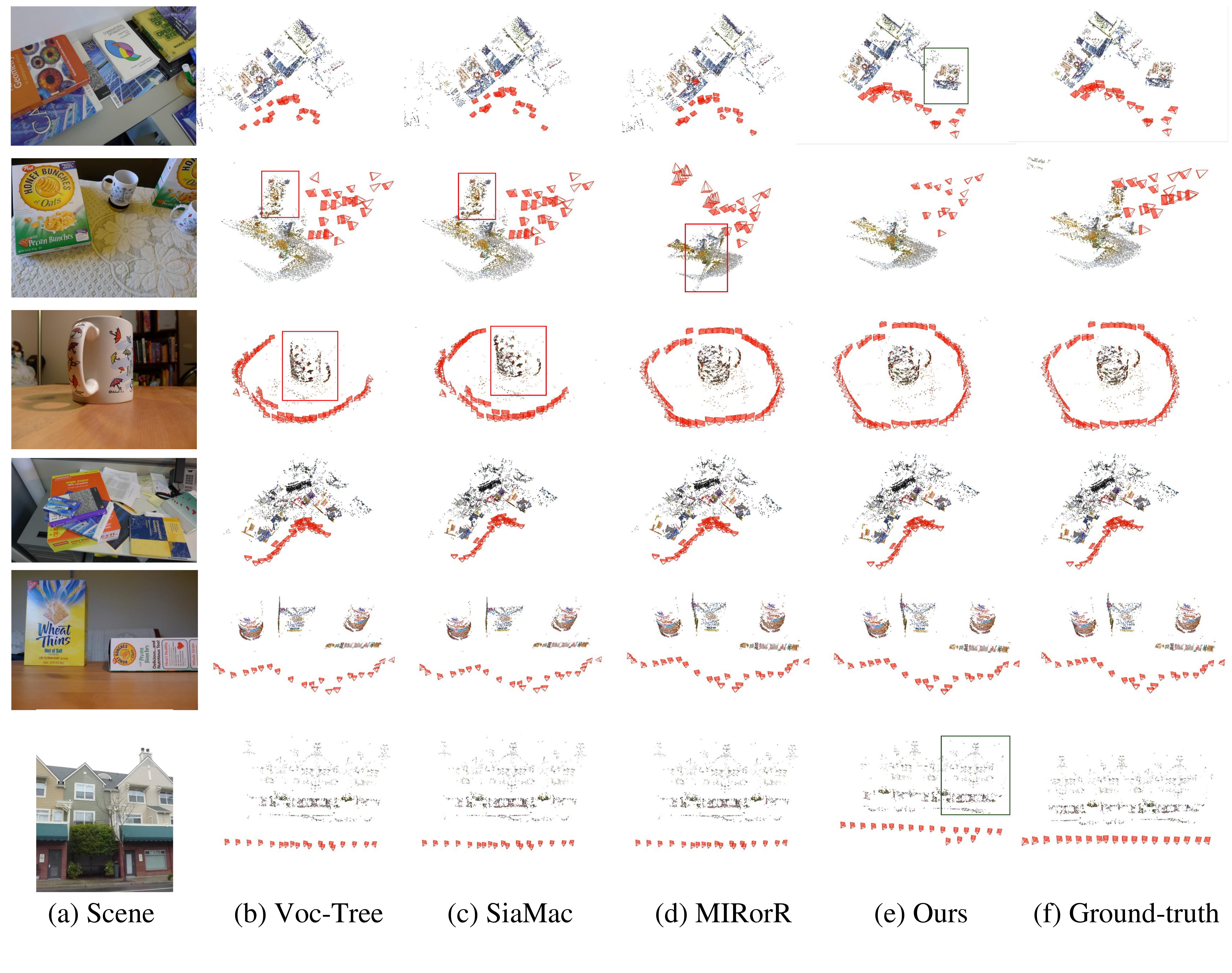}
	%\goodgap
	\caption{Experiment results on ambiguous dataset. From top to bottom are Books, Cereal, Cup, Desk, Oats and Street respectively. From left to right: the 1st column - one view of an ambiguous scene, from 2nd to 4th columns - SfM model using matches from Voc-Tree, SiaMAC and MIRorR, the 5th column - SfM model using our retrieved matches, the 6th column - SfM model using manually judged matches. Green box indicates structure only rebuilt by our method, while red box indicates wrong structure reconstructed by comparing methods.}
	\label{fig:ambiguous}
%	\vspace{-0.78cm}
\end{figure}

\begin{figure}
	\centering
	\subfigure[sensitivity of $(\tau_{MO},\tau_{CT})$ by MAC]{
		\label{fig:mac_ct_mo}
		\includegraphics[clip,trim=0mm 0mm 0mm 0mm,width=0.45\textwidth]{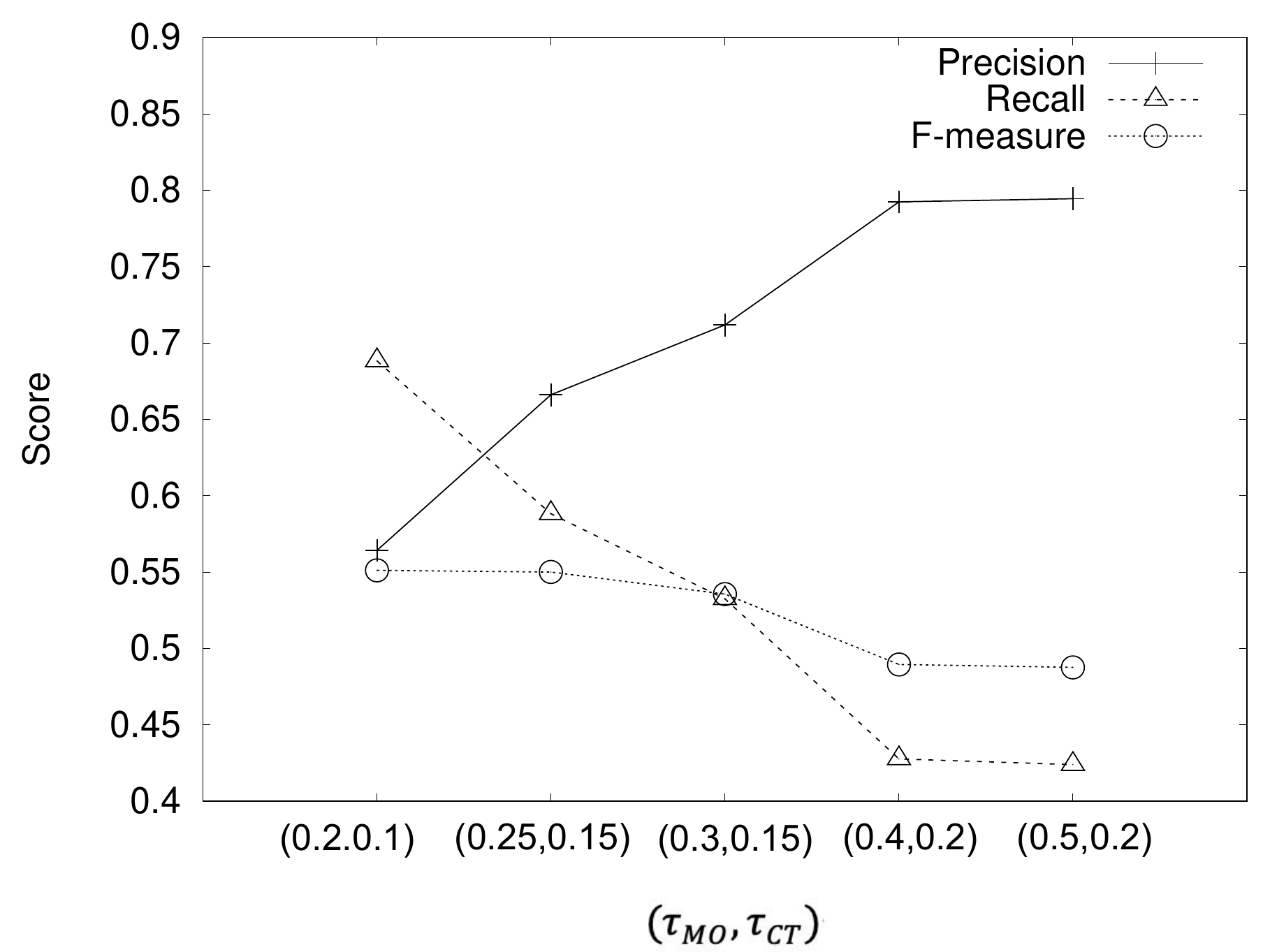}
	}
	%\goodgap
	\subfigure[sensitivity of $(\tau_{MO},\tau_{CT})$ by R-MAC]{
		\label{fig:rmac_ct_mo}
		\includegraphics[clip,trim=0mm 0mm 0mm 0mm,width=0.45\textwidth]{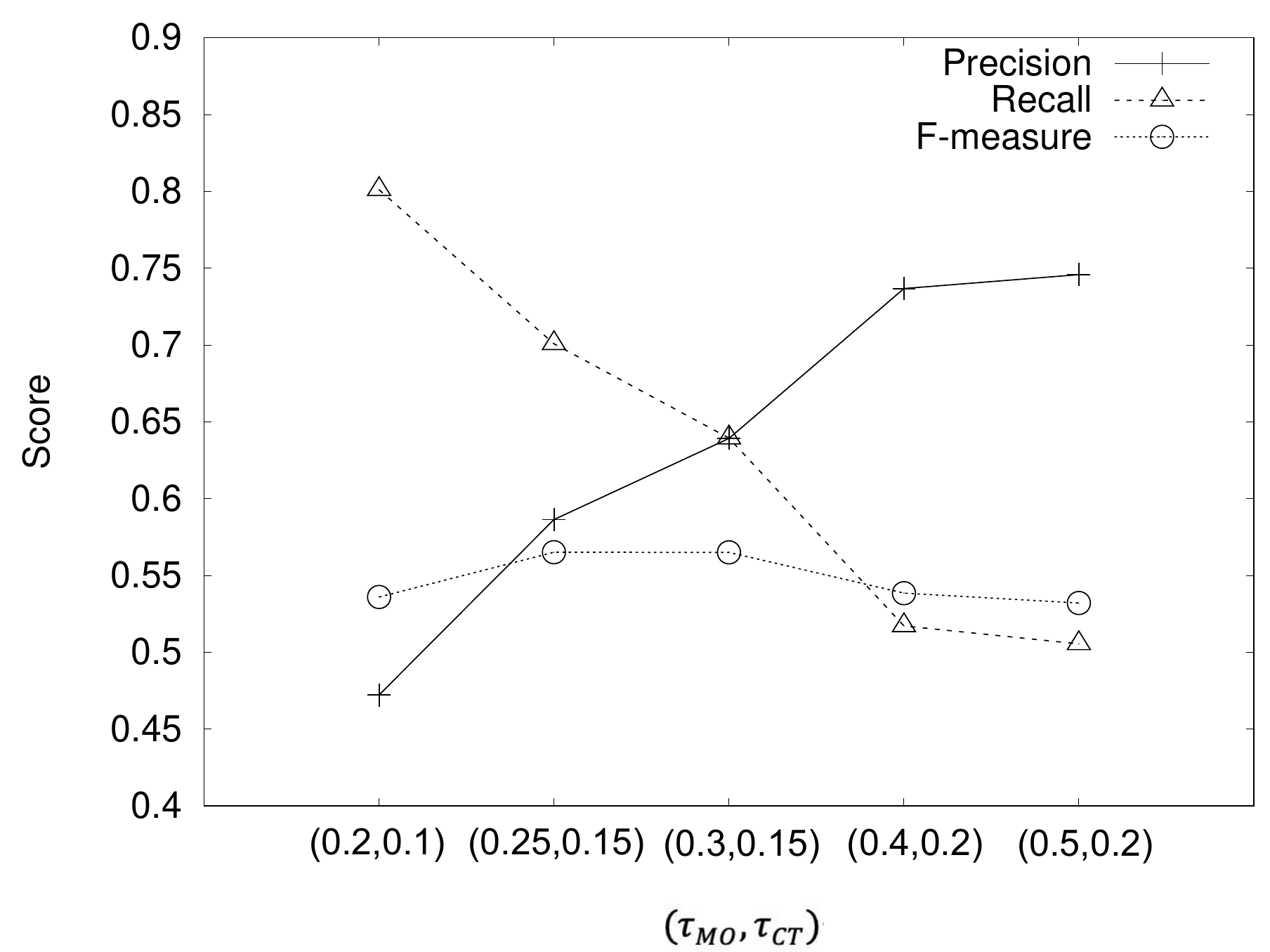}
	}
%    \vspace{-0.4cm}
	\caption{Sensitivity of $\tau_{MO}$, $\tau_{CT}$ and feature type.}
	\label{fig:para train}
%	\vspace{-0.78cm}
\end{figure}

To demonstrate the effect of GCN on image retrieval task, we use the view graph of Temple-of-Heaven to give an explanation in Fig~\ref{fig:effect}. In this dataset, the center part of view graph should be empty as front and back views cannot be matched. We can distinctly notice that our method has fewer false matches than pretrained CNN-based model.   

\begin{figure}
	\centering
	\subfigure[sensitivity of parameter $k_2$]{
		\label{fig:k2}
		\includegraphics[clip,trim=0mm 0mm 0mm 0mm,width=0.45\textwidth]{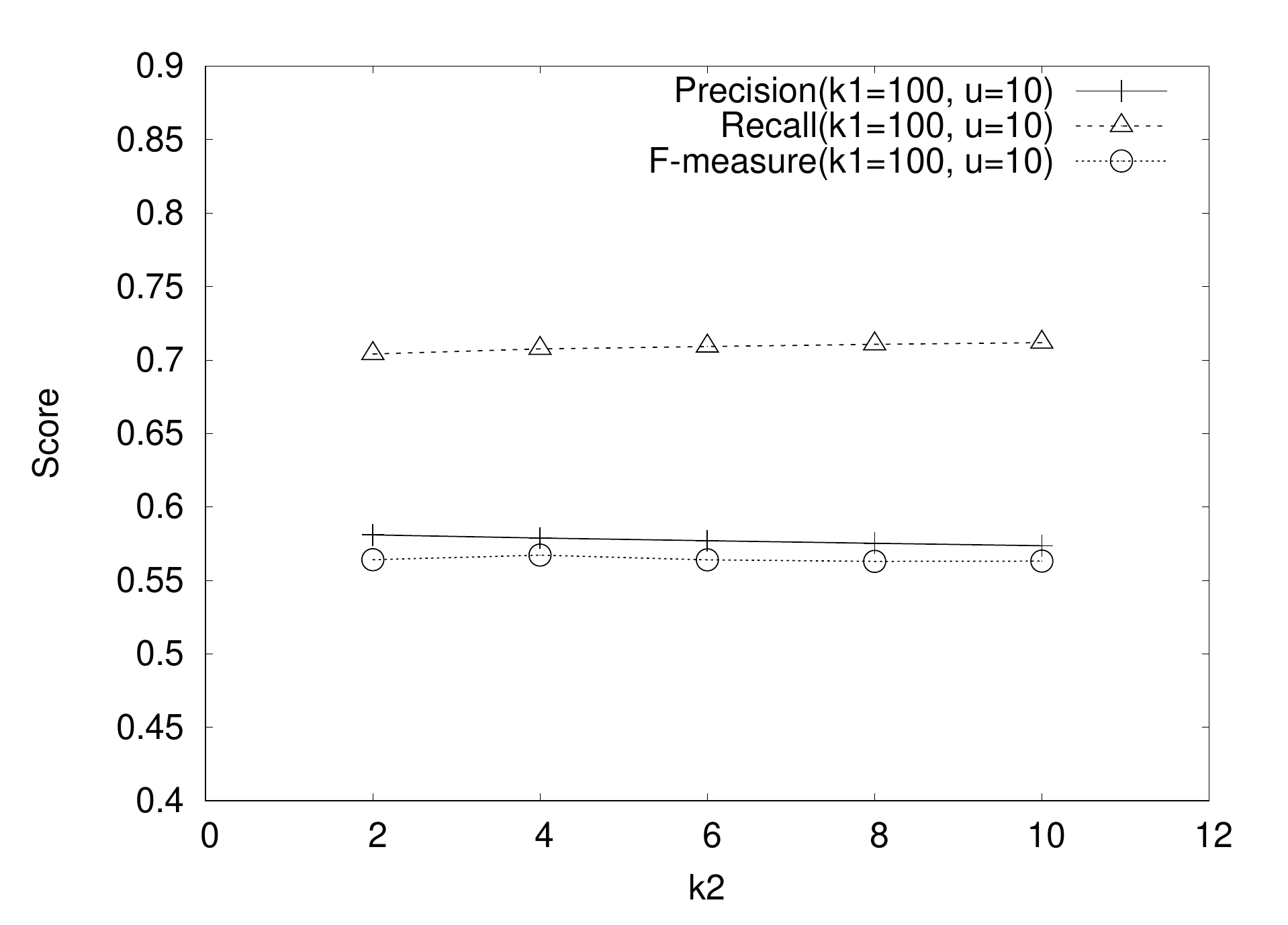}
	}
	%\goodgap
	\subfigure[sensitivity of parameter $k_1$]{
		\label{fig:k1}
		\includegraphics[clip,trim=0mm 0mm 0mm 0mm,width=0.45\textwidth]{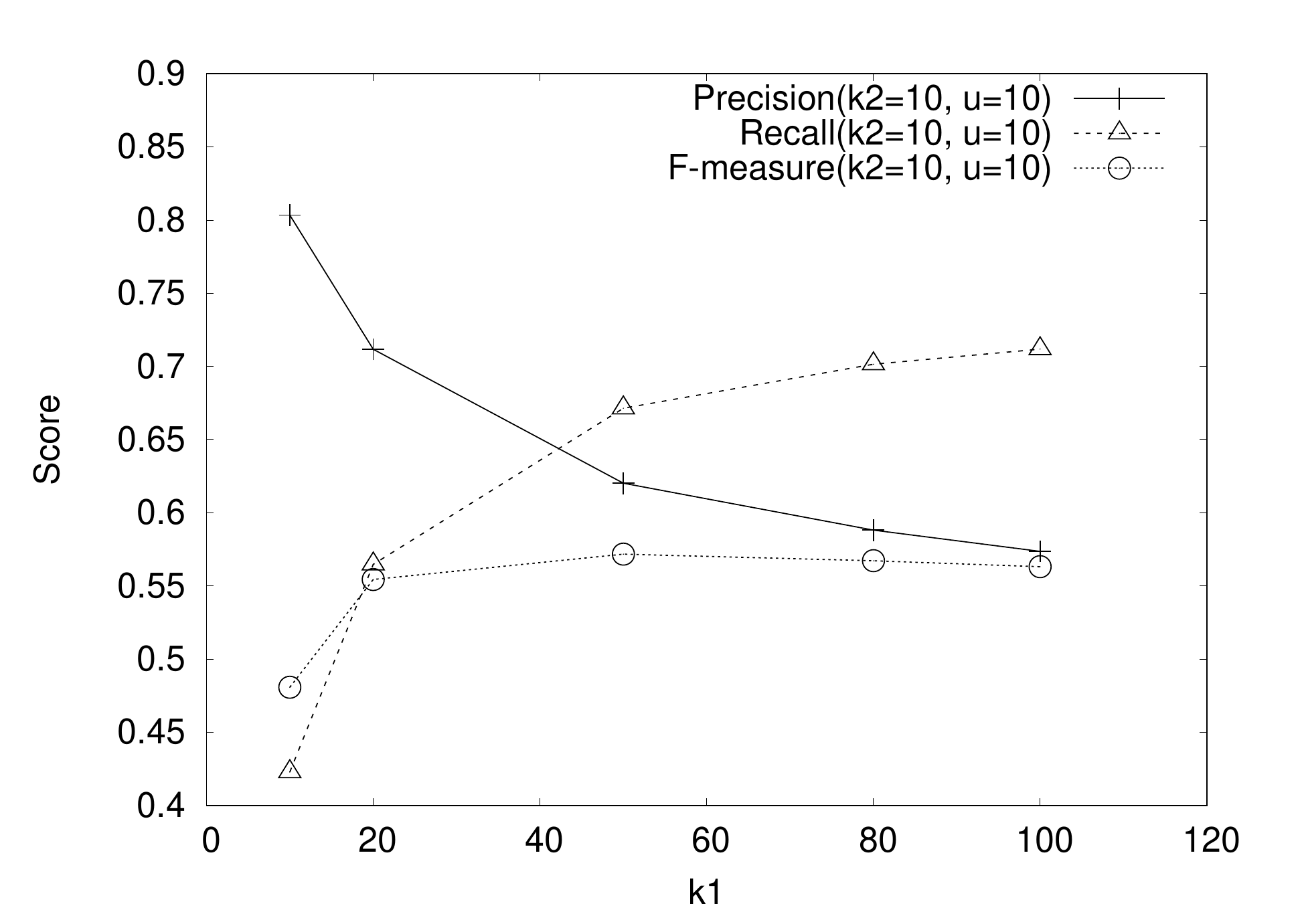}
	}
	%\goodgap
	\subfigure[sensitivity of parameter $u$ ($k_1 = 20$)]{
		\label{fig:u_k1_20}
		\includegraphics[clip,trim=0mm 0mm 0mm 0mm,width=0.45\textwidth]{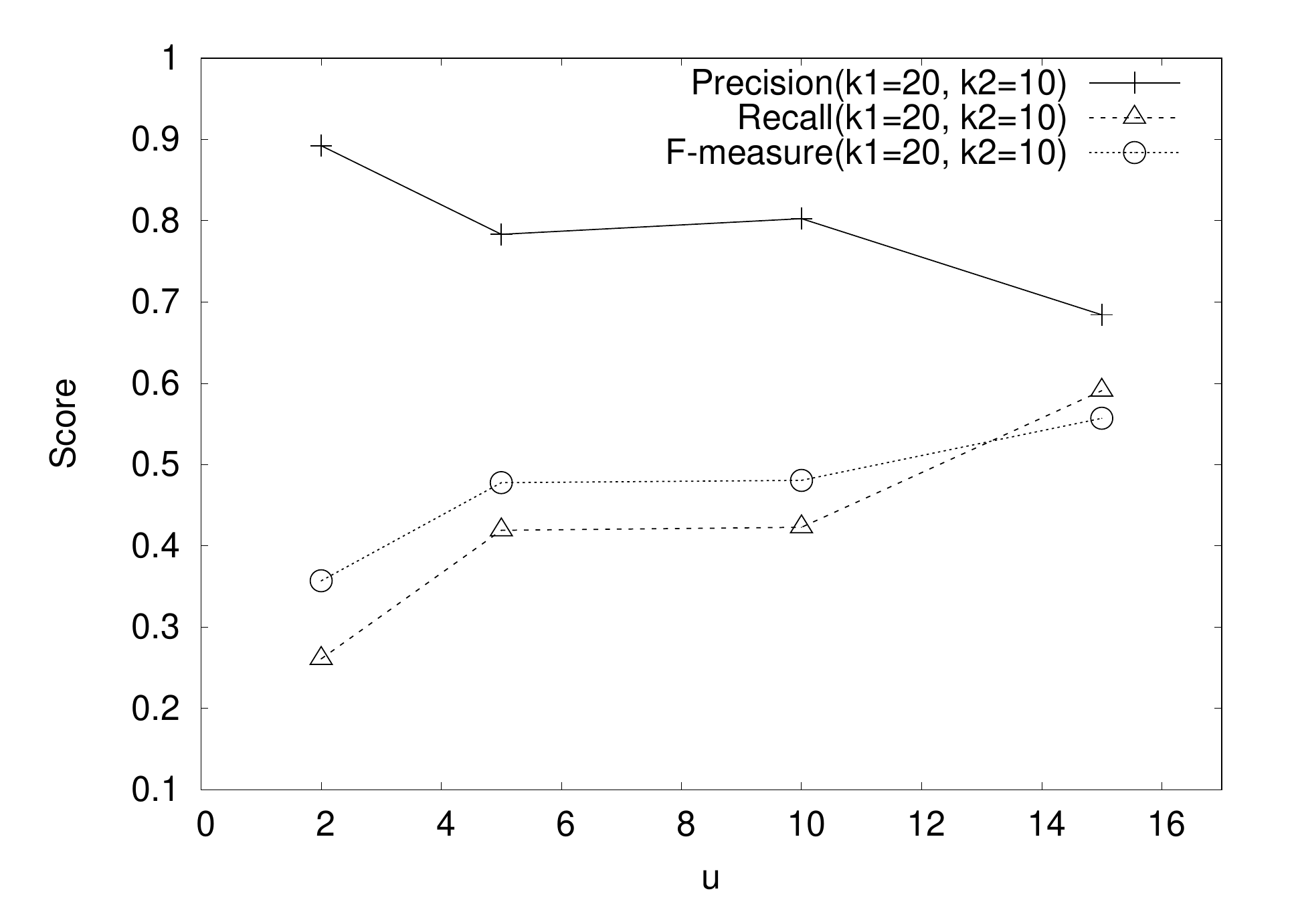}
	}
	%\goodgap
	\subfigure[sensitivity of parameter $u$ ($k_1 = 100$)]{
		\label{fig:u_k1_100}
		\includegraphics[clip,trim=0mm 0mm 0mm 0mm,width=0.45\textwidth]{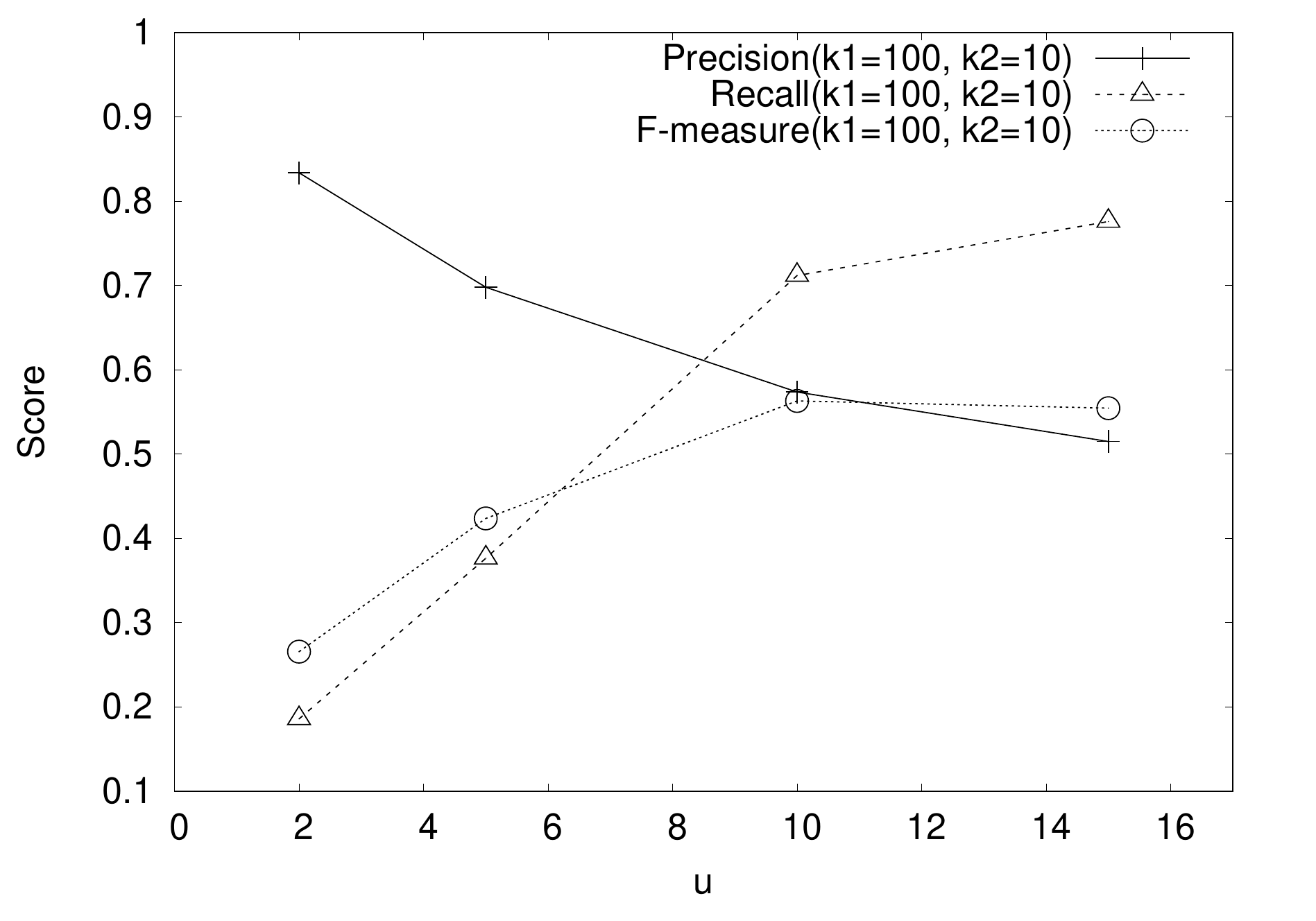}
	}
%    \vspace{-0.4cm}
	\caption{Sensitivity of parameters $k_1$, $k_2$ and $u$.}
	\label{fig:para test}
%	\vspace{-0.78cm}
\end{figure}

\section{Experiment}
\subsection{Datasets Overview}
To evaluate the effectiveness of our GCN-based image retrieval algorithm for SfM, we conduct experiments on different kinds of datasets: GL3D~\cite{shen2018matchable}, HKUST ambiguous dataset~\cite{shen2016graph}, public outdoor dataset~\cite{schonberger2016structure}, and 1DSfM dataset~\cite{wilson2014robust}.

GL3D is a large-scale dataset specially created for 3D reconstruction and geometry-related learning problems. GL3D provides the degree of mesh overlaps (${MO}_{ij}$) and common track(${CT}_{ij}$) between images pairs$(I_i,I_j)$ from accurate mesh re-projection, which serves to our supervised GCN training. HKUST ambiguous dataset is a small-scale dataset containing scenes with symmetric and duplicated structures. Public outdoor dataset includes medium-scale images specially for 3D reconstruction task. 1DSfM dataset consists of thousands of Internet photos downloaded from Flickr. Note that a large number of images in this dataset may be unrelated with 3D reconstruction.

\subsection{Evaluation Metrics}
For information retrieval system, mean Average Precision (mAP) is a very popular metric to measure the performance. However, mAP is not suitable for assessing image retrieval in SfM. There are two reasons for it: (1) as all retrieved items in the list (no matter ranked high or low) will be equally used for later SfM pipeline, the retrieval ranking is not important; (2) SfM pipeline has to confirm a retrieval number $k$ to form match pairs, the precision and recall calculation less than $k$ are meaningless. Therefore, we adopt precision, recall and F-measure with retrieval number $k$ to measure the experimental performance.

As for SfM evaluation, in matching procedure, we report the number of total attempted matches (TAM), the number of useful matches (UM) which pass geometric verification, and the running time to express the efficiency of our method. In mapping procedure, we record the number of recovered cameras, the number of sparse points, the number of observation points, and the re-projection error to describe the completeness and accuracy of rebuilt models.
\begin{table} [t]
	\small%/
	\centering
	\caption{Experimental results on the matchable image retrieval task (k=25).}
	\label{tab:image retrieve k25}
	\resizebox{\columnwidth}{!}{%
		\begin{tabular}{l@{~}|ccccccccc}
			\toprule
			&\multicolumn{1}{c}{VOC-Tree~\cite{shen2016graph}}&\multicolumn{3}{c}{SiaMAC~\cite{radenovic2016cnn}}&\multicolumn{3}{c}{MIRorR~\cite{shen2018matchable}}&\multicolumn{2}{c}{Ours}\\\cmidrule(r){2-2}\cmidrule(lr){3-5}\cmidrule(lr){6-8}\cmidrule(l){9-10}
			\ Score & depth=6, branch=8 & MAC & R-MAC& MAC + Lw &  MAC & R-MAC & PR-MAC &  MAC & R-MAC\\
			\midrule
			Precision      & 0.589 &  0.591 &0.5971 &0.6197 &0.6309&0.6263& 0.6422 & \textbf{0.6661} & 0.5864\\
			Recall          & 0.564 & 0.5626 &0.5667 &0.5936 &0.6015&0.5947& 0.6113 & 0.5882 & \textbf{0.7008}\\
			F-measure      & 0.498 & 0.4979 &0.5027 &0.5248 &0.5336&0.5288& 0.5434 &  0.55 & \textbf{0.5651}\\
			\bottomrule
		\end{tabular}%
	}
\end{table}

\begin{table} [t]
	\small%/
	\centering
	\caption{Experimental results on the matchable image retrieval task (k=100).}
	\label{tab:image retrieve k100}
	\resizebox{\columnwidth}{!}{%
		\begin{tabular}{l@{~}|ccccccccc}
			\toprule
			&\multicolumn{1}{c}{VOC-Tree~\cite{shen2016graph}}&\multicolumn{3}{c}{SiaMAC~\cite{radenovic2016cnn}}&\multicolumn{3}{c}{MIRorR~\cite{shen2018matchable}}&\multicolumn{2}{c}{Ours}\\\cmidrule(r){2-2}\cmidrule(lr){3-5}\cmidrule(lr){6-8}\cmidrule(l){9-10}
			\ Score & depth=6, branch=8 & MAC & R-MAC& MAC + Lw & MAC & R-MAC & PR-MAC &  MAC & R-MAC\\
			\midrule
			Precision      & 0.2288 &0.2712 &0.2726 &0.2835 & 0.2753&0.282& 0.2921   & \textbf{0.6661} & 0.5864\\
			Recall          & 0.7414 &0.8489 &0.8534 &0.8857 &0.8687&0.8792& \textbf{0.9027} &  0.5882 & 0.7008\\
			F-measure      & 0.3166  &0.3728 &0.3749 &0.3903 & 0.3801&0.3881& 0.4013 & 0.55 & \textbf{0.5651}\\
			\bottomrule
		\end{tabular}%
	}
\end{table}

\subsection{Baselines Introduction}
We include three baselines, VOC-Tree~\cite{shen2016graph}, MIRorR~\cite{shen2018matchable}, and SiaMAC~\cite{radenovic2016cnn}, which are used for image retrieval task especially in SfM. For a fair comparison, we tune the compared baselines to their best performance as described in their papers.

\subsection{Parameters Selection}
During QES construction, there are three kinds of hyper-parameters: $k_1$ and $k_2$ for discovering nodes; $u$ for appending edges; feature types MAC and R-MAC from pretrained embedding function~\cite{shen2018matchable} for calculating node features. Type MAC means that the representation features are extracted from full images, while type R-MAC implies that representation features are generated from summing up regional features of multiple different scales.

During GCN training on QES, ${MO}_{ij}$ and ${CT}_{ij}$ determine whether image pairs should be matched. We treat image pairs matched as long as ${MO}_{ij} \geqslant \tau_{MO}$ or ${CT}_{ij} \geqslant \tau_{CT}$, where $\tau_{MO}$ and $\tau_{CT}$ are pre-set threshold scores. 

In training phase, as only a few of matches matter in SfM, we select $k_1 = 100$. In order to avoid QES being too complicate to affect the efficiency, we set $k_2 = 5$ and $u = 10$. To explore the impact of various values of $\tau_{MO}$ and $\tau_{CT}$ and different feature types of MAC and R-MAC, we conduct two groups of experiments and the results are shown in Fig~\ref{fig:para train}. We find that R-MAC always has a better performance than MAC, and GCN model could get the highest F-measure score when setting $\tau_{MO} = 0.25$, $\tau_{CT} = 0.15$. In following experiments, we decide to adopt R-MAC as feature type.

In testing phase, as there is no need to follow the same settings with the training phase, we carry out comprehensive experiments to investigate the impact of different values of $k_1$, $k_2$ and $u$. The results are reported in Fig~\ref{fig:para test}. First, we keep $k_1$ and $u$ constant, vary $k_2$ to show how statistics change. We observe in Fig~\ref{fig:k2} that $k_2$ has no significant effect on the results. Next, we fix $k_2$ and $u$, test the sensitivity of parameter $k_1$. As reported in Fig~\ref{fig:k1}, larger $k_1$ brings more candidate links to be predicted, thus yielding higher recall but lower precision. At last, for parameter $u$, we conduct two groups of experiments. In Fig~\ref{fig:u_k1_20} and Fig~\ref{fig:u_k1_100}, we observe that parameter $u$ has a similar effect with parameter $k_1$.

\subsection{Experiments for Matchable Image Retrieval}
From parameter selection part, we find that settings with $\tau_{MO} = 0.25$, $\tau_{CT} = 0.15$, $k_1 = 100$, $k_2 = 5$, $u = 10$ have a balanced performance between precision and recall, and can get the highest F-measure score. In this section, we apply this configuration for the matchable image retrieval experiments in GL3D. 

The results are shown in Table~\ref{tab:image retrieve k25} with $k=25$ and Table~\ref{tab:image retrieve k100} with $k=100$ respectively. Our approach outperforms other comparing methods in terms of F-measure score. Through Table~\ref{tab:image retrieve k25} and Table~\ref{tab:image retrieve k100} we can clearly realize how difficult it is to select a proper retrieve number $k$ for previous research that can guarantee both completeness and efficiency. Our approach does not need to set this annoying parameter.

%------------------------------------------------------------------------- 
\begin{table} [t]
	\small%/
	\centering
	\caption{Experimental results of image matching on public outdoor dataset.}
	\label{tab:3d matching k25}
	\resizebox{\columnwidth}{!}{%
		\begin{tabular}{l@{~}|ccccccccc}
			\toprule
			&\multicolumn{2}{c}{VOC-Tree~\cite{shen2016graph}}&\multicolumn{2}{c}{SiaMAC~\cite{radenovic2016cnn}}&\multicolumn{2}{c}{MIRorR~\cite{shen2018matchable}}&\multicolumn{3}{c}{Ours}\\\cmidrule(r){2-3}\cmidrule(lr){4-5}\cmidrule(lr){6-7}\cmidrule(l){8-10}
			\ Scene & UM/TAM & Time (min) & UM/TAM& Time (min) & UM/TAM & Time (min) & UM/TAM & Time (min) & Speedup\\
			\midrule
			fc      & 1939/2339 (0.83) &0.400 &Fail &Fail & 2090/2624 (0.80)&0.440& 1412/1458 (\textbf{0.97})   & 0.260 & x1.5\\
			stadium & 1619/2699 (0.60) &1.367 &1513/2658 (0.57) &1.459 &1330/2685 (0.50)&1.523& 939/1298 (\textbf{0.72}) &  0.604 & x2.3\\
			garrard-hall      & 1030/1520 (0.68) &0.829 &1018/1520 (0.67) &0.810 & 1044/1567 (0.67)&0.867& 502/546 (\textbf{0.92}) & 0.272 & x3.0\\
			south-building      & 1546/2079 (0.74)  &1.263 &1613/2043 (0.79) &1.322 & 1650/1996 (0.83)&1.287& 900/941 (\textbf{0.96}) & 0.569 & x2.2\\
			graham-hall      & 5898/10475 (0.56)  &5.902 &5965/8845 (0.67) &4.964 & 5831/8600 (0.68) &4.740& 4846/6635 (\textbf{0.73}) & 3.935 & x1.2\\
			person-hall      & 3645/5734 (0.64)  &3.484 &3168/5162 (0.61)&3.185 & 3165/4986 (0.63)&3.031& 2422/3608 (\textbf{0.67}) & 2.199 & x1.4\\
			\bottomrule
		\end{tabular}%
	}
\end{table}

\begin{table} [t]
	\small%/
	\centering
	\caption{Experimental results of mapping on public outdoor dataset.}
	\label{tab:3d mapping}
	\resizebox{\columnwidth}{!}{%
		\begin{tabular}{l@{~}ccccccc}
			\toprule
			Scene & Method & Images & Registered & Sparse Points & Observations & Track Length & Repro. Error\\ 
			\midrule
			fc & VOC-Tree &150 & 150 & 26513 & 140062 & 5.2828 & 0.4839\\
			% & VOC-Tree (k=100) & & 150 & 28150 & 147216 & 5.2297 & 0.555\\
			
			 & SiaMAC  & & Fail & Fail & Fail & Fail & Fail \\
			% & SiaMAC (k=100) & & 150 & 28604 & 148631 & 5.1962 & 0.5236 \\
			 & MIRorR  & & 150 & 26295 & 136650&5.1968&0.4837 \\
			% & MIRorR (k=100) & & 150 & 28598 & 148628&5.1972&0.5227 \\
			 & Ours  & &150 &24738 & 126795 & 5.1255 & 0.4745\\
			\midrule
			stadium & VOC-Tree  &157 & 157 & 84723 & 381464 & 4.5024 & 1.0471\\
			% & VOC-Tree (k=100) & & 157 & 86182 & 388272 & 4.5053 & 1.0519\\
			
			 & SiaMAC  & & 154 & 81387 & 366107 & 4.5 & 1.0373 \\
			% & SiaMAC (k=100) & & 157 & 85829 & 386960 & 4.5085 & 1.0529 \\
			 & MIRorR  & & 156 & 77175 & 345419&4.4758&1.0279 \\
			% & MIRorR (k=100) & & 157 & 85336 & 384368&4.51&1.0516 \\
			 & Ours  & &155 &72632 & 322246 & 4.4367 & 1.0101\\
			\midrule
			garrard-hall & VOC-Tree & 100 & 100 & 57081 & 331992 & 5.8161 & 1.024\\
			% & VOC-Tree (k=100) & & 100 & 57335 & 332598 & 5.8009 & 1.0276\\
			
			 & SiaMAC & & 100 & 56920 & 331489 & 5.8238 & 1.025 \\
			% & SiaMAC (k=100) & & 100 & 57283 & 332629 & 5.8067 & 1.0229 \\
			 & MIRorR & & 100 & 57047 & 331838&5.817&1.0248 \\
			% & MIRorR (k=100) & & 100 & 57349 & 332678&5.801&1.023 \\
			 & Ours  & &100 &55184 & 319800 & 5.7952	 & 0.9997\\
			\midrule
			south-building & VOC-Tree & 128 & 128 & 85599 & 514822 & 6.0145 & 0.5909\\
			% & VOC-Tree (k=100) & & 128 & 86708 & 517652 & 5.97 & 0.5911\\
			
			 & SiaMAC & & 128 & 85660 & 514248 & 6.0033 & 0.5905 \\
			% & SiaMAC (k=100) & & 128 & 86778 & 517897 & 5.9681 & 0.5921 \\
			 & MIRorR & & 128 & 85626 & 514552&6.0093&0.5911 \\
			% & MIRorR (k=100) & & 128 & 86790 & 517894&5.967&0.5919 \\
			 & Ours  & &128 &83630 & 501698 & 5.999	 & 0.5824\\
			\midrule
			graham-hall & VOC-Tree & 562 & 556 & 271255 & 1711980 & 6.3113 & 1.0869\\
			% & VOC-Tree (k=100) & & 559 & 276458 & 1762234 & 6.3743 & 1.1041\\
			
			 & SiaMAC & & 559 & 292897 & 1603609 & 5.475 & 1.0268 \\
			% & SiaMAC (k=100) & & 560 & 281827 & 1687967 & 5.9894 & 1.0763 \\
			 & MIRorR & & 560 & 287487 & 1623510&5.6473&1.0492 \\
			% & MIRorR (k=100) & & 561 & 274537 & 1733893&6.3157&1.0928 \\
			 & Ours  & &555 &261222 & 1560843 & 5.9751	 & 1.0343\\
			\midrule
			person-hall & VOC-Tree & 330& 330 & 200196 & 1406306 & 7.0247 & 1.156\\
			% & VOC-Tree (k=100) & & 330 & 203718 & 1426459 & 7.0021 & 1.1745\\
			
			 & SiaMAC & & 238 & 139566 & 975117 & 6.9868 & 1.0845 \\
			% & SiaMAC (k=100) & & 330 & 202956 & 1423284 & 7.0213 & 1.1707 \\
			 & MIRorR & & 330 & 198354 & 1389135&7.0033&1.141 \\
			% & MIRorR (k=100) & &330 &202256 & 1419948 & 7.0205		 & 1.169 \\
			 & Ours  & &330 &202641 & 1362836 & 6.7254		 & 1.108\\
			\bottomrule
		\end{tabular}%
	}
\end{table}

\begin{table} [t]
	\small%/
	\centering
	\caption{Experimental results of image matching on 1DSfM dataset.}
	\label{tab:1dsfm matching k100}
	\resizebox{\columnwidth}{!}{%
		\begin{tabular}{l@{~}|ccccccccc}
			\toprule
			&\multicolumn{2}{c}{VOC-Tree~\cite{shen2016graph}}&\multicolumn{2}{c}{SiaMAC~\cite{radenovic2016cnn}}&\multicolumn{2}{c}{MIRorR~\cite{shen2018matchable}}&\multicolumn{2}{c}{Ours}\\\cmidrule(r){2-3}\cmidrule(lr){4-5}\cmidrule(lr){6-7}\cmidrule(l){8-10}
			\ Scene & UM/TAM & Time (min) & UM/TAM& Time (min) & UM/TAM & Time (min) & UM/TAM & Time (min) & Speedup \\
			\midrule
			Alamo      & 32532/214257 (0.15) & 82.144&54665/209073 (0.26)&75.730 & 27629/160767 (0.17)&57.890& 21740/49604 (\textbf{0.43})  & 19.612 & x3.0\\
			Ellis Island & Fail & Fail &46699/183036 (0.26)&62.286 & Fail & Fail & 23227/52847 (\textbf{0.44}) &  20.007 & x2.3\\
			Gendarmenmarkt      & 31494/100432 (0.31) &40.043 &43558/104397 (0.42) &40.736 & 20018/82438 (0.24) &31.626& 19875/32856 (\textbf{0.60}) & 13.291 & x2.4\\
			Madrid Metropolis      & 13477/97553 (0.14)  &32.712 &23337/92380 (0.25)&28.725 & 9584/73410 (0.13)&21.389& 9632/26543 (\textbf{0.36}) & 7.129 & x3.0 \\
			Roman Forum      & 39711/168677 (0.24) &65.761 &65516/168650 (0.39) &65.510 & 30894/129521 (0.24) &49.525& 29228/47952 (\textbf{0.61}) & 18.317 & x2.7 \\
			Tower of London      & 17181/115211 (0.15)  &44.538 &28761/110655 (0.26) &41.054 & 12536/87786 (0.14)&31.879& 11452/27272 (\textbf{0.42}) & 11.153 & x2.9\\
			\bottomrule
		\end{tabular}%
	}
\end{table}

\begin{table} [t]
	\small%/
	\centering
	\caption{Experimental results of mapping on 1DSfM dataset.}
	\label{tab:1dsfm mapping}
	\resizebox{\columnwidth}{!}{%
		\begin{tabular}{l@{~}ccccccc}
			\toprule
			Scene & Method & Images & Registered & Sparse Points & Observations & Track Length & Repro. Error\\ 
			\midrule
			%Alamo & VOC-Tree (k=25) &2915 & 895 & 154316 & 1798836 & 11.6568 & 0.6193\\
			Alamo & VOC-Tree  & 2915& 938 & 166907 & 2003316 & 12.0026 & 0.6496\\
			
			% & SiaMAC (k=25) & & 916 & 153458 & 1788370 & 11.6538 & 0.6164 \\
			 & SiaMAC  & & 967 & 172913 & 2067066 & 11.9544 & 0.6606 \\
			% & MIRorR (k=25) & & 882 & 150339 & 1634318&10.8709&0.6014 \\
			 & MIRorR  & & 925 & 159171 & 1894934&11.905&0.6364 \\
			 & Ours  & &862 &146979 & 1767485 & 12.0254 & 0.6317\\
			\midrule
			%Ellis Island & VOC-Tree (k=25) &2587 & F & F & F & F & F\\
			Ellis Island & VOC-Tree &2587 & Fail & Fail & Fail & Fail & Fail\\
			
			% & SiaMAC (k=25) & & F & F & F & F & F \\
			 & SiaMAC  & & 410 & 106275 & 738273 & 6.947 & 0.8137 \\
			% & MIRorR (k=25) & & 333 & 74602 & 468578&6.281&0.7588 \\
			 & MIRorR  & & Fail & Fail & Fail &Fail&Fail \\
			 & Ours  & &326 &87070 & 606156 & 6.9617 & 0.8109\\
			\midrule
			%Gendarmenmarkt & VOC-Tree (k=25) & 1463 & 1015 & 186224 & 1200190 & 6.4449 & 0.6948\\
			Gendarmenmarkt & VOC-Tree & 1463& 1057 & 208925 & 1366892 & 6.5425 & 0.7185\\
			
			% & SiaMAC (k=25) & & 1028 & 180916 & 1124253 & 6.2142 & 0.692 \\
			 & SiaMAC & & 1054 & 209716 & 1358449 & 6.4776 & 0.7217 \\
			% & MIRorR (k=25) & & 975 & 156368 & 938739&6.0034&0.6758 \\
			 & MIRorR  & & 980 & 168822 & 1090269&6.4581&0.7205 \\
			 & Ours  & &997 &181568 & 1138802 & 6.272	 & 0.6926\\
			\midrule
			%Madrid Metropolis & VOC-Tree (k=25) & 1344 & 421 & 64473 & 464567 & 7.2056 & 0.596\\
			Madrid Metropolis & VOC-Tree & 1344 & 452 & 71281 & 518857 & 7.279 & 0.6228\\
			
			% & SiaMAC (k=25) & & 474 & 68215 & 479159 & 7.0242 & 0.5881 \\
			 & SiaMAC & & 503 & 77667 & 554452 & 7.1388 & 0.6186 \\
			% & MIRorR (k=25) & & 415 & 55517 & 372451&6.7088&0.5493 \\
			 & MIRorR & & 489 & 66476 & 464502&6.9875&0.5898 \\
			 & Ours  & &420 &61111 & 437402 & 7.1575	 & 0.6098\\
			\midrule
			%Roman Forum  & VOC-Tree (k=25) & 2364 & 1459 & 292005 & 2390787 & 8.1875 & 0.7132\\
			Roman Forum  & VOC-Tree & 2364 & 1641 & 332421 & 2769546 & 8.3314 & 0.7308\\
			
			% & SiaMAC (k=25) & & 1640 & 308185 & 2422202 & 7.8596 & 0.6994 \\
			 & SiaMAC & & 1677 & 348262 & 2888378 & 8.2937 & 0.7339 \\
			% & MIRorR (k=25) & & 1557 & 272832 & 2081399&7.6289&0.6822 \\
			 & MIRorR & & 1623 & 306893 & 2506663&8.1678&0.7117 \\
			 & Ours  & &1575 &296194 & 2405047 & 8.1198	 & 0.7105\\
			\midrule
			%Tower of London & VOC-Tree (k=25) &1567 & 680 & 151396 & 1236577 & 8.1678 & 0.5979\\
			Tower of London & VOC-Tree & 1567 & 742 & 166622 & 13883301 & 8.332 & 0.6152\\
			
			% & SiaMAC (k=25) & & 733 & 154287 & 1241640 & 8.0476 & 0.5954 \\
			 & SiaMAC & & 765 & 171273 & 1426670 & 8.3298 & 0.62198 \\
			% & MIRorR (k=25) & & 670 & 134516 & 1057713&7.8631&0.5832 \\
			 & MIRorR & &740 &151625 & 1252361 & 8.2596		 & 0.6014 \\
			 & Ours  & &570 &135843 & 1151801 & 8.4789		 & 0.6078\\
			\bottomrule
		\end{tabular}%
	}
\end{table}
\subsection{Experiments for SfM}
In this section, we conduct plenty of reconstruction experiments to demonstrate the integration of image retrieval techniques with SfM. All reconstructions are implemented in the framework of COLMAP~\cite{schonberger2016structure}, which can be found in the supplementary material.

First, we report SfM results on challenging HKUST ambiguous dataset. As a small number of wrong pairwise matches in this dataset may cause a failure of reconstruction, the accuracy of image retrieval is extremely important. For our GCN model, we decide to apply configurations with $\tau_{MO} = 0.5$, $\tau_{CT} = 0.2$, $k_1 = 100$, $k_2 = 5$, $u = 10$ in training phase and $k_1 = 10$, $k_2 = 5$, $u = 2$ in testing phase. For fairness consideration, we select $k=5$ for compared retrieval methods to improve their accuracy. Since different retrieval methods lead to great diversities in reconstruction results, we directly show the rebuilding models in Fig~\ref{fig:ambiguous}. We find that our GCN-based retrieval method displays obvious advantages comparing visually-based retrieval approaches.

Second, we conduct experiments on public outdoor dataset. As this dataset is created especially for 3D reconstruction, we adopt a normal configuration with $\tau_{MO} = 0.25$, $\tau_{CT} = 0.15$, $k_1 = 100$, $k_2 = 5$, $u = 10$ in training phase and $k_1 = 100$, $k_2 = 5$, $u = 5$ in testing phase. We select $k=25$ for comparing methods in consideration of both efficiency and completeness. Table~\ref{tab:3d matching k25} provides statistics of matching process, and Table~\ref{tab:3d mapping} provides statistics of mapping process. The matching and mapping experiments imply that although our method generates minimal attempted match pairs, almost all of them could pass epipolar geometry verification and contribute to accurate and complete 3D models. This property can help to save massive computation resources especially in large-scale dataset.

%\begin{table} [t]
%	\small%/
%	\centering
%	\caption{Experimental results on image matching (k=100).}
%	\label{tab:3d matching k100}
%	\resizebox{\columnwidth}{!}{%
%		\begin{tabular}{l@{~}|ccccccccc}
%			\toprule
%			&\multicolumn{2}{c}{VOC-Tree~\cite{shen2016graph}}&\multicolumn{2}{c}{SiaMAC~\cite{radenovic2016cnn}}&\multicolumn{2}{c}{MIRorR~\cite{shen2018matchable}}&\multicolumn{3}{c}{Ours}\\\cmidrule(r){2-3}\cmidrule(lr){4-5}\cmidrule(lr){6-7}\cmidrule(l){8-10}
%			\ Scene & UM/TM & Time (min) & UM/TM& Time (min) & UM/TM & Time (min) & UM/TM & Time (min) & Speedup\\
%			\midrule
%			fc      & 4918/8795 (0.56) &1.524 &5139/9302 (0.55) &1.518 & 4956/9021 (0.55) & 1.568& 1412/1458 (0.97)   & 0.260  & x5.8\\
%			stadium & 2494/9691 (0.26) &5.080 &2489/9412 (0.26) &5.066 &2392/9620 (0.25) & 5.342& 939/1298 (0.72) &  0.604 & x7.9\\
%			garrard-hall  & 1311/4950 (0.26) &2.610 &1285/4950 (0.26)&2.601 & 1314/4950 (0.27) & 2.6& 502/546 (0.92) & 0.272  & x9.6\\
%			south-building      & 2519/7417 (0.34)  &4.805 &2610/7337 (0.36) &4.817 & 2602/7371 (0.35) & 4.796&  900/941 (0.96) & 0.569 & x8.2\\
%			graham-hall      & 12660/40685 (0.34)  &23.981 &10879/35741 (0.30)& 22.022 & 10989/35026 (0.31) & 21.120&4846/6635 (0.73) & 3.935& x5.4\\
%			person-hall      & 6595/22589 (0.29)  &14.427 &6137/22453 (0.27)&13.671 & 5444/20153 (0.27) & 12.233& 2422/3608 (0.67) & 2.199 & x5.6\\
%			\bottomrule
%		\end{tabular}%
%	}
%\end{table}

At last, we conduct experiments on 1DSfM dataset. Since 1DSfM dataset is downloaded from the Flickr, a large number of images in the dataset have nothing to do with final reconstruction, which is quite different from our training dataset GL3D. To narrow down this gap, we try to retrieve more images to prevent missing necessary images. We set $\tau_{MO} = 0.25$, $\tau_{CT} = 0.15$, $k_1 = 100$, $k_2 = 5$, $u = 10$ in training phase and $k_1 = 100$, $k_2 = 5$, $u = 15$ in testing phase. For comparing methods, we set $k=100$ due to the same reason. The matching results are reported in Table~\ref{tab:1dsfm matching k100}, and the mapping results are reported in Table~\ref{tab:1dsfm mapping}. The experiments show that we achieve comparable results with other retrieval approaches costing less computation.

%\begin{table} [t]
%	\small%/
%	\centering
%	\caption{Experimental results on image matching (k=25).}
%	\label{tab:1dsfm matching k25}
%	\resizebox{\columnwidth}{!}{%
%		\begin{tabular}{l@{~}|cccccccc}
%			\toprule
%			&\multicolumn{2}{c}{VOC-Tree~\cite{shen2016graph}}&\multicolumn{2}{c}{SiaMAC~\cite{radenovic2016cnn}}&\multicolumn{2}{c}{MIRorR~\cite{shen2018matchable}}&\multicolumn{2}{c}{Ours}\\\cmidrule(r){2-3}\cmidrule(lr){4-5}\cmidrule(lr){6-7}\cmidrule(l){8-9}
%			\ Scene & UM/TM & Time (min) & UM/TM& Time (min) & UM/TM & Time (min) & UM/TM & Time (min) \\
%			\midrule
%			Alamo      & 13960/55315 (0.25) & 21.226&18974/55096 (0.34) &20.712 & 12657/43259 (0.29)&16.264& 21740/49604 (0.43)  & 19.612 \\
%			Ellis Island & 16418/47231 (0.35) &17.753 &18985/48038 (0.40) &17.172 &12971/37972 (0.34)&13.696& 23227/52847 (0.44) &  20.007 \\
%			Gendarmenmarkt      & 13099/25895 (0.51)  &9.375 &15344/26805 (0.57)&9.552 & 9423/22094 (0.43)&8.200& 19875/32856 (0.60) & 13.291 \\
%			Madrid Metropolis      & 6053/24480 (0.25)  &7.974 &9024/23711 (0.38)&7.487 & 5081/19441 (0.26)&5.880& 9632/26543 (0.36) & 7.129 \\
%			Roman Forum      & 18036/43571 (0.41) &16.513 &24266/42506 (0.57) &15.951 & 15611/34557 (0.45)&13.123& 29228/47952 (0.61) & 18.317 \\
%			Tower of London      & 8049/29239 (0.28)  &11.671 &11531/28453 (0.41) &10.775 & 6596/22831 (0.29)&8.589& 11452/27272 (0.42) & 11.153 \\
%			\bottomrule
%		\end{tabular}%
%	}
%\end{table}

\section{Conclusion}
In this paper, we propose a novel image retrieve method for SfM via Graph Convolutional Network (GCN). We emphasize that the local context surrounding query image provides rich information about the matchable likelihood between this image and its neighbors. By constructing Query Enclosing Subgraph (QES) to depict the local context, we adopt GCN to directly predict whether test image pairs share scene overlaps. Extensive experiments indicate that the proposed method can handle challenging scenes with ambiguous structure, and significantly reduces the number of image pairs for matching without degrading the quality of subsequent SfM pipeline.

%------------------------------------------------------------------------- 

%\subsection{Citations}

%References are listed in alphabetic order by the surname of the first author, or the identifying word (e.g., in case of a website). Have
%all anonymized references at the beginning of the list.

%here would be your acknowledgement (if any) in the final accepted paper

%===========================================================
\bibliographystyle{splncs}
\bibliography{egbib}

%this would normally be the end of your paper, but you may also have an appendix
%within the given limit of number of pages
\end{document}